\documentclass[acmsmall]{acmart}
\usepackage[utf8]{inputenc}
\usepackage{graphicx}

\usepackage{array}
\usepackage{tabularx}
\usepackage{longtable}
\usepackage{booktabs}
\usepackage{makecell}
\usepackage{float}
\usepackage[table]{xcolor}   
\usepackage{ragged2e}
\usepackage{colortbl}        

\newcolumntype{L}[1]{>{\RaggedRight\arraybackslash}p{#1}}

\usepackage{tikz}
\usetikzlibrary{arrows.meta,positioning,calc}

\usepackage{listings}

\colorlet{punct}{red!60!black}
\definecolor{delim}{RGB}{20,105,176}
\colorlet{numb}{magenta!60!black}

\lstdefinelanguage{json}{
  basicstyle=\footnotesize\ttfamily,
  numbers=none,
  showstringspaces=false,
  breaklines=true,
  morecomment=[l]{//},
  literate=
   *{0}{{{\color{numb}0}}}{1}
    {1}{{{\color{numb}1}}}{1}
    {2}{{{\color{numb}2}}}{1}
    {3}{{{\color{numb}3}}}{1}
    {4}{{{\color{numb}4}}}{1}
    {5}{{{\color{numb}5}}}{1}
    {6}{{{\color{numb}6}}}{1}
    {7}{{{\color{numb}7}}}{1}
    {8}{{{\color{numb}8}}}{1}
    {9}{{{\color{numb}9}}}{1}
    {:}{{{\color{punct}{:}}}}{1}
    {,}{{{\color{punct}{,}}}}{1}
    {\{}{{{\color{delim}{\{}}}}{1}
    {\}}{{{\color{delim}{\}}}}}{1}
    {[}{{{\color{delim}{[}}}}{1}
    {]}{{{\color{delim}{]}}}}{1}
}

\usepackage[most]{tcolorbox}

\definecolor{tinymlblue}{RGB}{70,172,200}

\newtcolorbox[auto counter]{codebox}[2][]{%
  float=htb,
  floatplacement=htb,
  colback=blue!2,
  colframe=blue!15,
  coltitle=black,
  fonttitle=\bfseries,
  fontupper=\footnotesize,
  title=Box~\thetcbcounter: #2,
  label={#1}
}

\newtcolorbox{hardwarebox}[1]{%
  colback      = white,
  colframe     = tinymlblue,
  coltitle     = white,
  colbacktitle = tinymlblue,
  fonttitle    = \bfseries,
  title        = {\textit{#1}},
  boxrule      = 0.9pt,
  arc          = 2pt,
  left         = 1.25em,
  right        = 1.25em,
  top          = 0.9em,
  bottom       = 0.9em,
}

\DeclareUnicodeCharacter{202F}{\,}

\AtBeginDocument{%
  }

\setcopyright{acmlicensed}
\copyrightyear{2026}
\acmYear{2026}
\acmDOI{TBD}

\acmJournal{TAAS}

\begin{document}

\title{Externalizing Context for Autonomous and Adaptive Systems: A Survey of MCP-Inspired Architectures and Evaluation}

\author{Gaurab Chhetri}
\email{gaurab@txstate.edu}
\orcid{0009-0000-0124-4814}
\affiliation{%
  \institution{Texas State University}
  \city{San Marcos}
  \state{Texas}
  \country{USA}
}

\author{Shriyank Somvanshi}
\orcid{0009-0008-3723-0607}
\email{shriyank@txstate.edu}
\affiliation{%
  \institution{Texas State University}
  \city{San Marcos}
  \state{Texas}
  \country{USA}
}

\author{Md Monzurul Islam}
\orcid{0009-0007-3670-6100}
\email{monzurul@txstate.edu}
\affiliation{%
  \institution{Texas State University}
  \city{San Marcos}
  \state{Texas}
  \country{USA}
}

\author{Shamyo Brotee}
\orcid{0009-0008-4094-3536}
\email{nmx23@txstate.edu}
\affiliation{%
  \institution{Texas State University}
  \city{San Marcos}
  \state{Texas}
  \country{USA}
}

\author{Mahmuda Sultana Mimi}
\orcid{0009-0007-8534-3633}
\email{qnb9@txstate.edu}
\affiliation{%
  \institution{Texas State University}
  \city{San Marcos}
  \state{Texas}
  \country{USA}
}

\author{Dipti Koirala}
\email{yfn21@txstate.edu}
\affiliation{%
  \institution{Texas State University}
  \city{San Marcos}
  \state{Texas}
  \country{USA}
}

\author{Biplov Pandey}
\orcid{0009-0003-2166-9387}
\email{iub14@txstate.edu}
\affiliation{%
  \institution{Texas State University}
  \city{San Marcos}
  \state{Texas}
  \country{USA}
}

\author{Subasish Das, Ph.D.}
\orcid{0000-0002-1671-2753}
\email{subasish@txstate.edu}
\affiliation{%
  \institution{Texas State University}
  \city{San Marcos}
  \state{Texas}
  \country{USA}
}

\renewcommand{\shortauthors}{Chhetri et al.}

\begin{abstract}

Autonomous and adaptive systems increasingly close control loops across distributed sensing, learning, and actuation, yet many deployments still optimize protocol layers, middleware, and applications in isolation. This fragmentation limits situational awareness, produces conflicting adaptation decisions, and weakens coordination under uncertainty. This survey examines the Model Context Protocol (MCP) as a unifying substrate for context sharing and adaptation in distributed autonomous systems. We analyze how MCP’s session semantics, structured context exchange, and capability negotiation can externalize context and adaptation state, supporting shared world models and more consistent coordination among autonomous agents. Synthesizing insights from adaptive transport protocols, context-aware middleware, and multi-agent autonomy, we introduce a taxonomy that organizes MCP-inspired designs by where adaptation logic resides and how context is represented and exchanged. Using this taxonomy, we compare MCP-enabled patterns with established approaches and identify design trade-offs in scalability, robustness, and controllability. We conclude with an evaluation agenda for autonomy-centric assessment, emphasizing adaptation latency, coordination consistency, and safety under partial observability, and discuss transport systems as a representative domain with implications for robotics, IoT, and edge AI.

\end{abstract}



\received{March 1, 2026}

\maketitle

\section{Introduction}

\subsection{Fragmentation in Autonomous and Adaptive Systems}

Autonomous and adaptive systems increasingly rely on distributed sensing, learning, and actuation to close control loops across heterogeneous environments, including autonomous vehicles, edge--cloud infrastructures, and large-scale Internet of Things (IoT) deployments. Despite this complexity, many real-world systems still optimize protocol stacks, middleware, and applications in isolation, a limitation that becomes acute in dynamic and safety-critical settings \cite{bonaventure2015beyondtcp, ericsson2018quic}. Transport protocols such as TCP and UDP were designed for relatively stable conditions and now operate in environments where topology, workload, and safety constraints evolve rapidly \cite{zou2006adaptive, pejovic2011context}. Consequently, adaptation decisions made at one layer often conflict with those made elsewhere. This fragmentation manifests in three dimensions. 
First, \emph{protocol fragmentation} arises when heterogeneous transports such as QUIC, vehicular V2X protocols, and lightweight IoT mechanisms adapt independently based on local signals \cite{ericsson2018quic, talkingiot2025protocolwars}. Second, \emph{context fragmentation} occurs when rich telemetry, sensor outputs, and application state remain siloed, preventing coherent system-wide reasoning \cite{du2024survey, li2015context}. Third, \emph{decision fragmentation} results when control policies are implemented independently across layers and subsystems, producing conflicting adaptations and inefficient resource allocation \cite{cordis101117675, tudelft_adapt_or}. In safety-critical domains such as autonomous transportation, incomplete context sharing between perception, communication, and control layers can degrade safety and robustness. More broadly, the absence of standardized mechanisms for sharing context and adaptation state limits distributed coordination under uncertainty \cite{alanazi2024framework, russo2012unifying}.

\subsection{Why MCP Matters}

Model Context Protocol (MCP) has emerged as a candidate approach to mitigating this fragmentation by standardizing how context, capabilities, and interaction state are represented and exchanged \cite{anthropic2024mcp}. Although originally introduced to support integration between AI models and external tools, MCP embodies architectural principles aligned with distributed autonomy: persistent sessions, structured JSON-RPC messaging, and explicit capability negotiation. In this survey, we treat MCP not as a transport replacement but as an \emph{adaptation substrate} that externalizes context and adaptation state so that heterogeneous agents can participate in shared sense--reason--act loops. By exposing contextual information and available actions through standardized interfaces, MCP enables broader and more consistent world models than isolated adaptation mechanisms allow. Consider an urban intersection involving autonomous vehicles, traffic signals, pedestrian sensing infrastructure, and emergency services. In fragmented systems, each entity adapts independently. In contrast, an MCP-enabled environment allows standardized exchange of contextual information such as traffic density, pedestrian intent forecasts, network conditions, and urgency signals, supporting coordinated adaptation across agents \cite{ray2025survey}. While transport systems provide a safety-critical example, the underlying coordination challenge extends to robotics, IoT, and edge AI systems more broadly.

\subsection{Research Gap}

Existing research addresses transport protocols, context-aware computing, and integration architectures largely as separate threads. Transport surveys emphasize congestion control and reliability \cite{he2005survey, velayutham2005transport}; context-aware frameworks focus on acquisition and reasoning \cite{li2015context, baldauf2007survey}; integration studies examine cross-layer optimization and unified control \cite{russo2012unifying, cordis101117675}. However, limited work systematically analyzes how structured context exchange mechanisms such as MCP influence closed-loop adaptation across system boundaries. This gap is timely. MCP remains under active evolution, and autonomous systems increasingly require principled coordination across heterogeneous components. An autonomy-centric analysis can inform both MCP development and evaluation in safety-critical environments.

\subsection{Contributions}

This survey develops an autonomy-oriented perspective on MCP as a coordination and context-exchange layer for distributed adaptive systems. Our contributions are:

\begin{enumerate}
    \item We position MCP’s session model, structured context representations, and capability negotiation mechanisms as a shared world-model interface for distributed sense--reason--act loops.
    \item We introduce a five-category taxonomy comprising Adaptive Protocol Mechanisms, Context-Aware Frameworks, Unification Models, Transport System Integration, and MCP-Enabled Architectures, and reinterpret them as loci of adaptation.
    \item We map representative systems and prior surveys onto this taxonomy to analyze how MCP-style context exchange mitigates protocol, context, and decision fragmentation.
    \item We propose an autonomy-centric evaluation agenda, including metrics such as adaptation latency, coordination consistency, decision conflict rate, and recovery under partial observability.
    \item We provide an experimentation foundation through the curated repository \textbf{\textit{awesome-mcp}}\footnote{\url{https://github.com/gauravfs-14/awesome-mcp}}, aggregating tools, libraries, and resources for MCP-based autonomy research.
\end{enumerate}

\subsection{Organization}

\S~\ref{sec:related_work} reviews adaptive transport, interoperability foundations, and context-aware systems. 
\S~\ref{sec:taxonomy} presents the proposed taxonomy. 
\S~\ref{sec:mcp} and \S~\ref{sec:protocols} summarize MCP architecture and protocol principles. 
\S~\ref{sec:context} examines MCP-enabled context coordination scenarios. 
\S~\ref{sec:evaluation} discusses evaluation methodologies. 
\S~\ref{sec:challenges} outlines open challenges, and \S~\ref{sec:future} highlights research directions. 
\S~\ref{sec:conclusion} concludes.

\section{Background and Positioning}
\label{sec:related_work}

Adaptive transport systems integrate artificial intelligence, real-time analytics, and heterogeneous connectivity to respond dynamically to evolving traffic, user demands, and environmental constraints \cite{maadi2022real, calabro2023adaptive, roman2024model, farooq2024adaptive}. Such systems increasingly require structured context exchange rather than isolated protocol optimizations. From a communication perspective, traditional transports such as TCP and UDP provide reliable delivery, while QUIC introduces reduced latency, multiplexing, and integrated security \cite{mansour2021quic, kyratzisQuic}. Multipath communication and adaptive congestion control improve robustness under heterogeneous link conditions \cite{chao2021brief, barre2011multipath}. For MCP, the central requirement is not replacing these transports but enabling semantically rich context exchange atop them. Context-aware computing provides the conceptual foundation for MCP. Architectures span distributed, centralized, and layered designs \cite{li2015context, baldauf2007survey, du2024survey}. Ontology-based modeling using OWL/RDF supports semantic interoperability \cite{krummenacher2007analyzing, villalonga2016ontology, guermah2014ontology}, while hybrid reasoning approaches combine rules and machine learning to manage uncertainty \cite{ranganathan2003middleware}. Within Multi-Agent Systems, these capabilities yield context-aware agents capable of sensing, reasoning, and acting in dynamic environments \cite{du2024survey}. Quality of Context (QoC) concepts such as trustworthiness and timeliness \cite{huebscher2004adaptive} directly inform MCP’s context selection and routing logic. Prior surveys emphasize individual strands of this space. Transport-focused surveys analyze congestion control and reliability \cite{he2005survey, velayutham2005transport}; context-aware surveys catalog middleware and modeling approaches \cite{baldauf2007survey, li2015context, du2024survey}; integration studies explore cross-layer optimization \cite{russo2012unifying, cordis101117675}; and emerging MCP overviews describe architectural features \cite{hou2025model, krishnan2025advancing, ray2025survey, schmid2025mcp}. In contrast, this survey anchors MCP explicitly within adaptive transport and autonomous system coordination, clarifying where MCP complements or extends prior unification approaches.

\section{Taxonomy of Unifying Models for Adaptive Transport Systems
}
\label{sec:taxonomy}

\begin{figure}[htbp]
  \centering
  \includegraphics[width=0.95\linewidth]{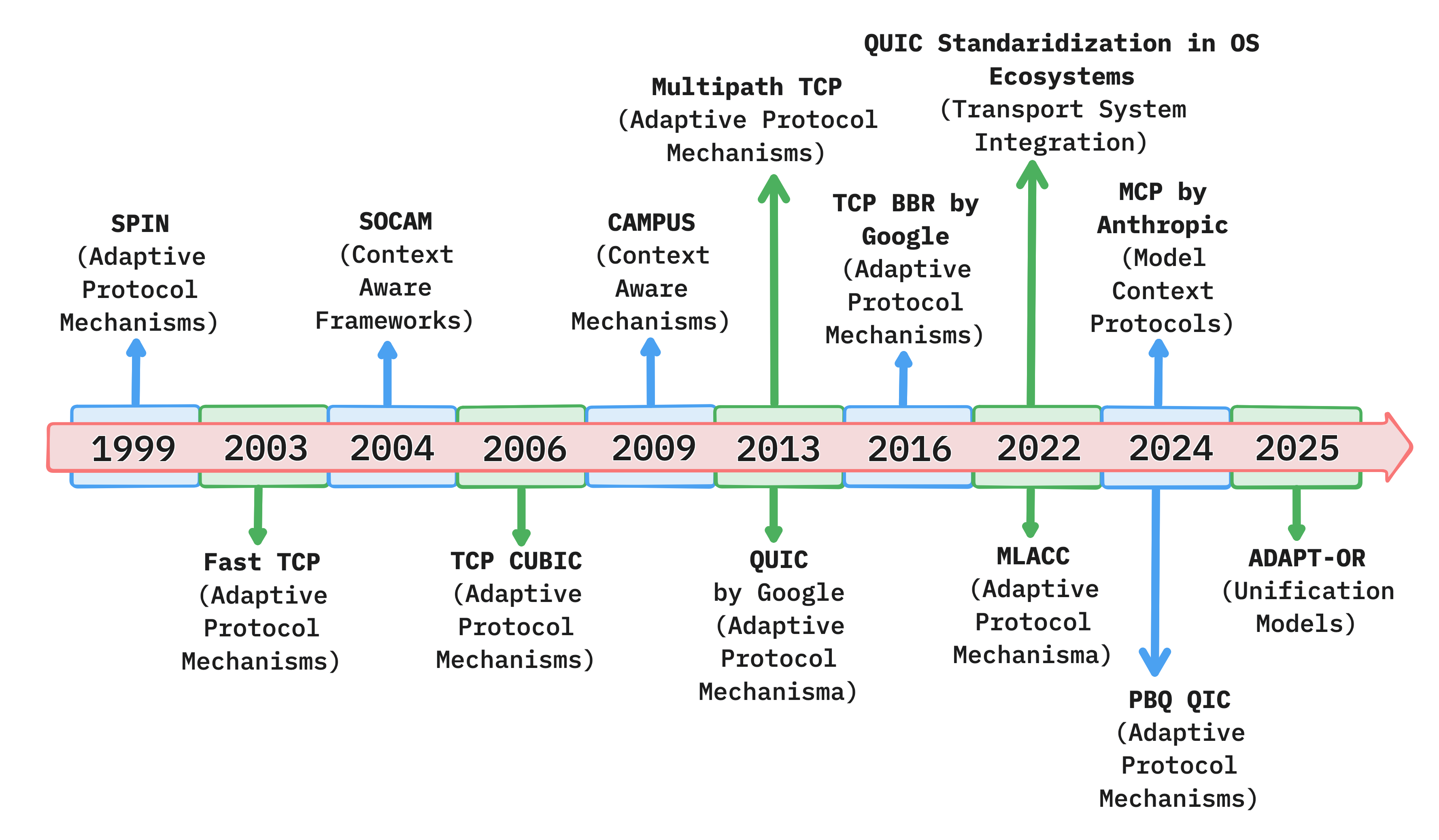}
  \caption{Comprehensive taxonomy and timeline of unifying models for adaptive transport systems, showing the five primary categories and their subcategories.}
  \label{fig:TaxonomyPic}
\end{figure}

This section develops a taxonomy of unifying models for adaptive transport systems organized into five categories (see~\autoref{fig:TaxonomyPic}): \textit{Adaptive Protocol Mechanisms}, \textit{Context-Aware Frameworks}, \textit{Unification Models}, \textit{Transport System Integration}, and \textit{Model Context Protocols}. We interpret these as loci of adaptation in distributed autonomy: local, reactive protocol behaviors; perception and state-estimation layers; control-loop composition; distributed actuation across infrastructures; and MCP-enabled coordination and policy sharing. The taxonomy is derived from a systematic review of research and real-world implementations, building on prior surveys and classification studies. In line with calls for integrated frameworks that can bridge the diverse components of modern mobility systems \cite{tudelft_adapt_or, cordis101117675}, it offers a structured view of how adaptation, context management, and interoperability can be combined into cohesive designs. To avoid conflating abstraction levels, we treat \textit{Adaptive Protocol Mechanisms} and \textit{Context-Aware Frameworks} as component-level techniques in sense–reason–act loops, \textit{Unification Models} and \textit{Transport System Integration} as architectural arrangements that connect multiple loops, and \textit{Model Context Protocols} as architectures that instantiate MCP on top of those layers to expose shared adaptation state across agents; some systems naturally span multiple categories, and we explicitly note such overlaps in the mapping tables that follow.

\begin{table}[htbp]
\centering
\small
\caption{Representative works mapped to taxonomy categories.}
\label{tab:taxonomy_mapping}
\renewcommand{\arraystretch}{1.2}
{\fontsize{8}{9}\selectfont
\begin{tabularx}{\textwidth}{L{2.5cm} L{2cm} X}
\toprule
\rowcolor{blue!6}
\textbf{Category} & \textbf{Representative Works} & \textbf{Primary mechanisms / transport examples} \\
\midrule
Adaptive Protocol Mechanisms &
\cite{he2005survey, elbery2023toward, zhang2023pbq} &
Congestion-, flow-, and error-control adaptations for heterogeneous links and highly variable traffic patterns; our taxonomy uses these works as exemplars of protocol-layer adaptation, not as evidence of convergence toward MCP. \\
\midrule
Context-Aware Frameworks &
\cite{li2015context, baldauf2007survey, du2024survey} &
Middleware and sensing frameworks for acquiring, modeling, and reasoning over user, network, and environmental context in ITS deployments; we treat MCP as one possible consumer and producer of context within such frameworks. \\
\midrule
Unification Models &
\cite{russo2012unifying, elatec2025smartmoves} &
Architectural patterns that combine multiple adaptive services and layers into coherent control loops for mobility and communication systems; these works predate MCP and provide baselines against which we compare MCP-enabled designs. \\
\midrule
Transport System Integration &
\cite{alanazi2024framework, talkingiot2025protocolwars, sung2023decentralized} &
Frameworks that embed adaptive transport mechanisms into cloud/edge, IoT, and SDN-based platforms supporting vehicular and multimodal mobility, illustrating integration concerns that any MCP deployment would need to respect. \\
\midrule
MCP-Enabled Architectures &
\cite{hou2025model, krishnan2025advancing, ray2025survey, schmid2025mcp} &
Emerging designs that instantiate MCP as a context-exchange and tool-orchestration layer above existing transport, ITS, and middleware components; these are treated as early examples rather than endpoints toward which all prior systems were implicitly evolving. \\
\bottomrule
\end{tabularx}
}
\end{table}

Several systems in Table~\ref{tab:taxonomy_mapping} legitimately occupy more than one category—for example, large-scale ITS platforms often combine adaptive congestion control (Adaptive Protocol Mechanisms), context brokers (Context-Aware Frameworks), and cross-layer service orchestration (Unification Models). Rather than forcing a single label or implying a trajectory toward MCP, we use the taxonomy to clarify which aspect of a system is under discussion at a given point in the survey.

\subsection{Adaptive Protocol Mechanisms}

Adaptive Protocol Mechanisms comprise techniques that allow transport protocols to modify behavior in response to changing network, application, and environmental conditions \cite{velayutham2005transport, he2005survey}. Rather than cataloguing all congestion-control variants or VANET routing schemes, we treat this body of work as evidence that transport protocols already embed rich adaptation logic—adjusting sending rates, error recovery, and QoS handling to changing conditions \cite{he2005survey, elbery2023toward, zhang2023pbq}. Our focus is on how such mechanisms can be exposed and coordinated through higher-level unification models and, ultimately, through MCP-style context exchange.

Beyond congestion control, adaptive mechanisms address flow control optimization, aligning transmission rates with network capacity and receiver processing capabilities \cite{atkin2003evaluation, pejovic2011context}. Predictive and context-aware models adjust flow windows or pacing rates in real time \cite{atkin2003evaluation, anvari2023adaptive}. Error recovery has evolved from simple retransmission to approaches that account for error causes, application tolerances, and network characteristics \cite{velayutham2005transport, zou2006adaptive, pejovic2011context}. Bandwidth adaptation adjusts transmission rates reactively and proactively using forecasts of available capacity \cite{atkin2003evaluation, pejovic2011context}, while QoS management ensures that critical or latency-sensitive applications receive priority treatment \cite{velayutham2005transport, zou2006adaptive}. Together, these mechanisms provide the operational foundation on which higher-level context-aware and unifying frameworks build.

\subsection{Context-Aware Frameworks}

Context-aware frameworks provide the infrastructure for acquiring, interpreting, and using contextual information to inform adaptation strategies within transport systems \cite{devx2025contextaware, pejovic2011context}. They are indispensable in multi-agent and distributed environments where timely, accurate context underpins decisions for autonomous vehicles, dynamic traffic signal control, and large-scale mobility-as-a-service platforms \cite{du2024survey, li2015context}. Typical functions include environmental context processing, which monitors network conditions, device capabilities, and operational constraints \cite{jevinger2023context, laicontext}; application context recognition, which captures performance goals, priorities, and user-defined preferences \cite{linqing2023dynamic, pejovic2011context}; and user behavior adaptation, which exploits historical interaction patterns to optimize future behavior \cite{linqing2023dynamic, anvari2023adaptive}. Effective frameworks combine distributed network-state awareness with resource-availability monitoring so that adaptation decisions reflect current conditions \cite{jevinger2023context, devx2025contextaware, kumar2003dynamic, cordis101117675}. To limit monitoring overhead, systems employ sampling, predictive modeling, or selective reporting while preserving accuracy \cite{zou2006adaptive, govloop2020adaptivenetwork}. By linking low-level sensing to high-level decision-making, context-aware frameworks bridge raw data and actionable intelligence, enabling more targeted adaptive protocols.

\subsection{Unification Models}

While adaptive protocols and context-aware frameworks provide essential components, Unification Models supply the architectural strategies for integrating them into cohesive systems \cite{russo2012unifying, elatec2025smartmoves}. Protocol-layer abstraction maintains consistent interfaces even as underlying transports change with network conditions \cite{velayutham2005transport, iren1999transport}. Service integration models combine congestion control, error recovery, and context processing into unified frameworks that address complex application requirements \cite{elatec2025smartmoves, cordis101117675}. Cross-layer optimization breaks down traditional boundaries between protocol layers, enabling coordinated decisions that optimize performance across the stack \cite{pejovic2011context, govloop2020adaptivenetwork}, but must be balanced against modularity and maintainability. Standardization and interoperability solutions address the alignment of adaptation mechanisms and context representations \cite{hamptons2025standardization, wikipedia2025commprotocol, talkingiot2025protocolwars}, often relying on protocol translation, context mapping, and negotiation to ensure compatibility across vendors and domains.

\subsection{Transport System Integration Architectures}

Transport System Integration concerns embedding adaptive transport mechanisms within broader computing and communication ecosystems \cite{alanazi2024framework}. In multi-modal environments, this requires coordination between protocols operating over different network types and physical media \cite{russo2012unifying, cordis101117675}. IoT deployments extend these demands, as transports must accommodate constrained devices, intermittent connectivity, and heterogeneous capabilities \cite{talkingiot2025protocolwars, alanazi2024framework}. Integration with edge computing allows localized processing of adaptation logic, reducing latency and conserving bandwidth by minimizing centralized computation \cite{bhardwaj2020adaptive, alanazi2024framework}. Cloud-native protocols exploit elasticity and dynamic resource allocation to handle auto-scaling, service migration, and fluctuating workloads \cite{alanazi2024framework, sung2023decentralized}. Together, these strategies ensure that adaptive mechanisms function effectively within diverse and evolving environments.

\section{
Model Context Protocol (MCP): Architecture, Framework, and Mechanisms
}
\label{sec:mcp}

\begin{figure}
    \centering
    \includegraphics[width=\linewidth]{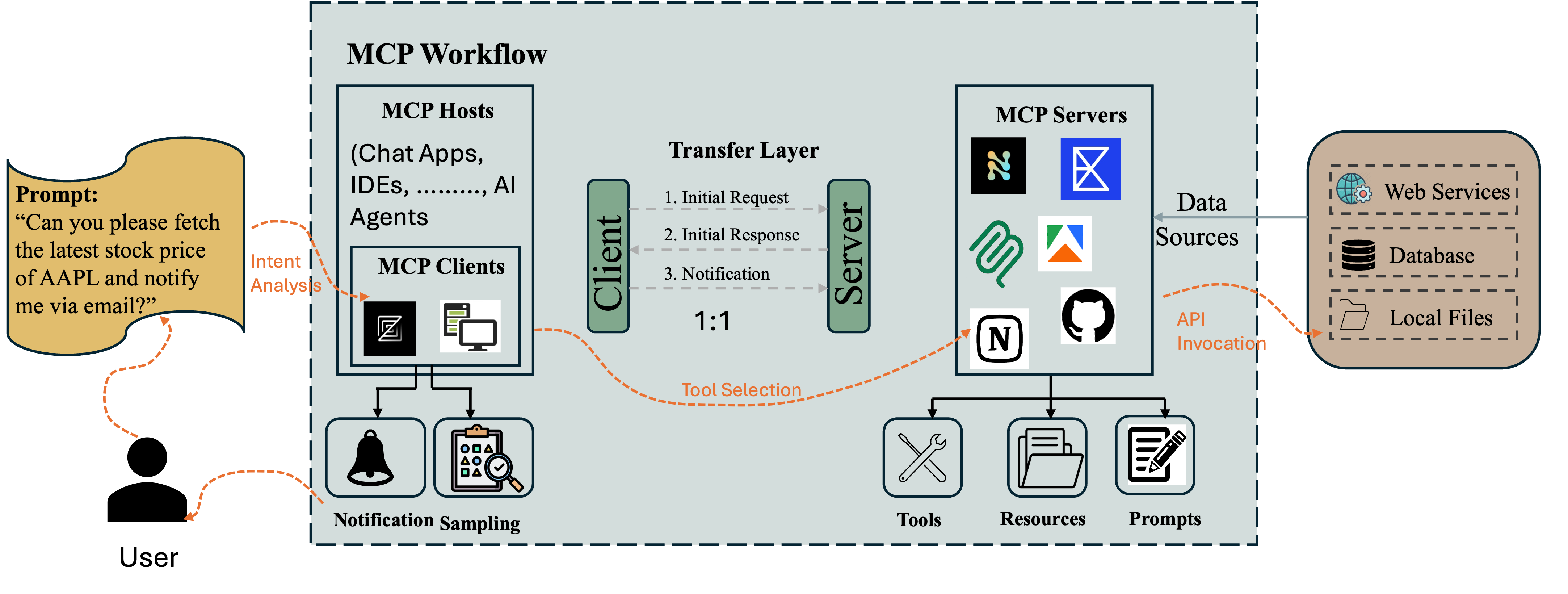}
    \caption{MCP Architecture and Workflow}
    \label{fig:mcp-arch}

\end{figure}

The MCP is an open framework for secure, structured integration between AI-powered systems and external data sources. Developers can expose data via MCP servers or build AI applications (MCP clients) that connect to these servers to consume contextual data \cite{anthropic2024mcp}. As shown in figure~\ref{fig:mcp-arch}, this architecture supports secure two-way communication between data sources and AI tools. Together with the specification summary in \S~\ref{sec:protocols}, this description reflects MCP as it exists today, based on public documentation and early implementations.
Beyond its basic client–server architecture, MCP defines a full-stack framework with data-ingestion specifications, contextual metadata tagging, and interoperability standards across platforms. It plays a role in multi-tool agent workflows, allowing AI systems to coordinate operations such as combining document search with messaging APIs to support reasoning across distributed systems. The remainder of this section interprets these mechanisms through the lens of adaptive transport systems; unless explicitly noted as deployed, the architectures and scenarios should be understood as prospective applications rather than existing MCP deployments.

\subsection{Context Representation Models}
Context representation is fundamental to MCP. It specifies how information such as user identity, device attributes, environmental state, and application requirements is structured, interpreted, and exchanged between systems. MCP supports this through standardized schemas and semantics that formalize context so that systems can reason about dynamic environments consistently \cite{contextmodel2016, 5483753}. Representations are extensible rather than domain-locked: lightweight models such as ContextML \cite{5483753} encode context as categorized areas (e.g., user, time, location) and integrate into broader workflows via REST-based communication. As data shifts from static documents to dynamically assembled, context-sensitive content \cite{content2018}, MCP helps synchronize AI behavior with real-time situational changes by ensuring interoperable tagging, transformation, and ingestion across desktop agents, API clients, and distributed services \cite{anthropic2024mcp,schmid2025mcp}.

Handling real-world contexts remains challenging. Modeling uncertainty, temporal dependencies, and multi-dimensional relationships demands semantic expressiveness and computational scalability. MCP’s ontology-aware metadata structure can improve interoperability and reuse \cite{5206920}, but inconsistencies arise when formal models do not match collected metadata, particularly in domains dominated by implicit or location-based context \cite{5206920}. Conventional modeling languages often struggle with variant-heavy or fluid data, limiting applicability in content-rich settings such as adaptive UIs or personalized recommendation systems \cite{content2018}. Maintaining persistent, bidirectional connections between clients and servers also introduces protocol overhead \cite{anthropic2024mcp}. As context-aware AI evolves, future MCP implementations will need more adaptive modeling layers, finer-grained ontological reasoning, and better scalability across edge and cloud infrastructures.

\subsection{Context Exchange Mechanisms}

Building on the JSON-RPC context exchange summarized in \S\ref{sec:protocols} and illustrated in Box~\ref{box:json}, the MCP literature explores how these mechanisms support real-time and asynchronous tasks, multi-transport communication, and capability negotiation. We therefore focus on deployment-relevant aspects such as transport choices and feature advertisement.

MCP context sharing supports metadata ranging from environmental signals and system configurations to user-specific attributes. Developers exchange this information in structured formats that preserve semantic meaning while enabling interoperability \cite{ray2025survey}. Clients can advertise supported features (for example file access or messaging) and negotiate agreements to coordinate behavior \cite{ray2025survey}. A dual-transport design lets implementers choose stdio for embedded settings or Streamable HTTP for event-driven communication in distributed systems \cite{ray2025survey, ehtesham2025surveyagentinteroperabilityprotocols}. These options can facilitate flexible context exchange and tool coordination, but also raise questions about synchronization, overhead, and alignment with evolving privacy standards \cite{anthropic2024mcp, 5483753}.

\subsection{Contextual Decision Making}

Contextual decision making in MCP relies on context modeling techniques that enable systems to adjust dynamically to evolving environments \cite{contextmodel2016}. Through standardized mechanisms for ingesting, transforming, and tagging metadata, MCP provides decision pipelines with access to rich information drawn from diverse domains \cite{schmid2025mcp}. These pipelines run on top of session and context-exchange mechanisms summarized in \S\ref{sec:protocols}, and exploit multi-tool workflows in which contextual reasoning integrates specialized resources to deliver coordinated, near-real-time responses \cite{5483753}.

Machine learning models increasingly populate these pipelines, learning adaptation strategies from historical context patterns and refining decisions as data evolves \cite{contextmodel2016, ray2025survey}. This shift from static rules to learning-based decision making is a central development in context-aware computing. MCP work further suggests that contextual decision making benefits from multi-objective optimization that balances performance, resource efficiency, and security guarantees \cite{contextmodel2016, krishnan2025advancing}. In multi-agent environments, standardized context exchange supports coordination so that agents can account for both local and global priorities \cite{krishnan2025advancing}, while online learning methods improve adaptability by updating strategies without interrupting operations \cite{contextmodel2016, ray2025survey}.

Viewed through an autonomy lens, MCP-based pipelines support policy selection over shared context state, multi-objective optimization across safety, efficiency, and resource constraints, and decision consistency among agents that coordinate under partial and asynchronous observations. Experiences from open-source ML collaborations indicate additional benefits, such as shared improvements in parameters, model structures, and validation tools, which may strengthen reliability and cross-domain applicability \cite{BhatiaEtAl2023}. Traditional integration approaches—point-to-point adapters, domain-specific middleware, service-oriented architectures, and message brokers—have all aided interoperability but face limits in scalability, semantic expressiveness, and dynamic coordination. Table~\ref{tab:integration_comparison} contrasts these approaches with MCP, highlighting both historical roles and MCP’s distinctive characteristics.

\begin{codebox}[box:json]{JSON-RPC Structure}
\footnotesize

\begin{minipage}[t]{0.48\linewidth}
\texttt{// 1. Route-planning request from an in-vehicle MCP client}
\begin{lstlisting}[language=json]
{
  "jsonrpc": "2.0",
  "id": "req-route-1",
  "method": "tools/routes/plan",
  "params": {
    "origin": {"lat": 29.8851, "lon": -97.9405},
    "destination": {"lat": 29.8893, "lon": -97.9421},
    "constraints": {
      "max_travel_time_s": 600,
      "avoid_links": ["RSU-23-RSU-24"],
      "priority": "safety"
    }
  }
}
\end{lstlisting}

\texttt{// 2. Success response from a routes tool}
\begin{lstlisting}[language=json]
{
  "jsonrpc": "2.0",
  "id": "req-route-1",
  "result": {
    "route_id": "r-4821",
    "expected_travel_time_s": 540,
    "steps": [
      {"segment": "RSU-10-RSU-11", "distance_m": 820},
      {"segment": "RSU-11-RSU-15", "distance_m": 650}
    ],
    "context_version": "net-topology:v3.2"
  }
}
\end{lstlisting}
\end{minipage}\hfill
\begin{minipage}[t]{0.48\linewidth}
\texttt{// 3. Error response when a policy blocks changes}
\begin{lstlisting}[language=json]
{
  "jsonrpc": "2.0",
  "id": "req-route-2",
  "error": {
    "code": 403,
    "message": "Forbidden: route changes disabled in school zone",
    "data": {"zone_id": "school-17"}
  }
}
\end{lstlisting}

\texttt{// 4. Notification from a network-telemetry MCP tool}
\begin{lstlisting}[language=json]
{
  "jsonrpc": "2.0",
  "method": "events/network/link_state_changed",
  "params": {
    "link_id": "RSU-23-RSU-24",
    "new_state": "degraded",
    "expected_latency_increase_ms": 30,
    "timestamp": "2025-01-01T08:23:10Z"
  }
}
\end{lstlisting}
\end{minipage}

\end{codebox}

\begin{table}[htbp]
\centering
\small
\caption{Comparison of legacy integration approaches and Model Context Protocol (MCP).}
\label{tab:integration_comparison}
\renewcommand{\arraystretch}{1.2}
{\fontsize{8}{9}\selectfont
\begin{tabularx}{\textwidth}{L{2cm}L{3cm}L{2.8cm}L{3cm}X}
\toprule
\rowcolor{blue!6}
\textbf{Approach} & \textbf{Characteristics} & \textbf{Limitations} & \textbf{Relevance to MCP} & \textbf{References} \\
\midrule
Point-to-Point Adapters &
Custom connectors between specific systems or applications. &
High maintenance cost; poor scalability as the number of systems grows. &
MCP replaces ad hoc adapters with a standardized, protocol-level interface. &
\cite{du2024survey} \\
\midrule
Domain-Specific Middleware &
Middleware tailored to specific domains (e.g., IoT, vehicular networks). &
Limited cross-domain applicability; requires domain expertise for extension. &
MCP offers domain-agnostic extensibility with semantic interoperability. &
\cite{li2015context, baldauf2007survey} \\
\midrule
Service-Oriented Architectures (SOA) / REST APIs &
Expose functionality through standardized HTTP/REST services. &
Syntactic interoperability only; lacks semantic context exchange; limited capability negotiation. &
MCP introduces semantically rich JSON-RPC messages and dynamic capability discovery. &
\cite{baldauf2007survey} \\
\midrule
Message Brokers (e.g., MQTT, AMQP) &
Support publish–subscribe communication for distributed systems. &
Efficient delivery but weak semantic guarantees; limited context sharing beyond payload. &
MCP builds on message-passing but standardizes context semantics and tool orchestration. &
\cite{du2024survey} \\
\midrule
Model Context Protocol (MCP) &
Protocol-level standard for context representation, exchange, and orchestration. &
Still emerging; adoption and standardization challenges remain. &
Provides unified, extensible integration with semantic and context-awareness capabilities. &
\cite{anthropic2024mcp, hou2025model, krishnan2025advancing} \\
\bottomrule
\end{tabularx}
}
\end{table}

As Table~\ref{tab:integration_comparison} shows, earlier methods primarily addressed syntactic interoperability or domain-specific challenges, often at the expense of generality and scalability. We therefore treat MCP as one integration option alongside REST/HTTP APIs, message brokers (e.g., MQTT, AMQP), and domain-specific middleware used in ITS and IoT platforms. In settings with strict real-time or certification requirements these alternatives may remain preferable, while MCP’s value lies where semantic context exchange, capability negotiation, and cross-domain tool orchestration justify its additional complexity; throughout the paper we compare MCP-enabled architectures against such baselines rather than assuming MCP as the default unifying layer.

\section{Core Principles of MCP}
\label{sec:protocols}

The Model Context Protocol (MCP) is specified as an open, JSON-RPC–based framework for secure, structured, and interoperable context exchange between autonomous agents, AI-powered services, and external tools \cite{anthropic2024mcp, hou2025model}. We summarize MCP as it exists today—drawing primarily from its public specification and early implementations—before considering how these mechanisms apply to adaptive transport and other cyber-physical domains. MCP operates through a client–server paradigm in which developers expose contextual data via MCP servers or consume such data through MCP clients \cite{anthropic2024mcp}. Structured context representation models allow diverse information—ranging from environmental telemetry to policy constraints and learned value functions—to be encoded and interpreted consistently across systems \cite{schmid2025mcp}, while context exchange mechanisms define protocols and interfaces for distributing that information across heterogeneous systems \cite{anthropic2024mcp}. Contextual decision-making frameworks then use this information to inform adaptation strategies under real-time constraints \cite{hou2025model, linqing2023dynamic, schmid2025mcp, laicontext}. We therefore view MCP as a coordination layer for autonomous control loops—one that can unify adaptive protocols, context-aware frameworks, and integration architectures into a coherent ecosystem for transport and other safety-critical systems.

MCP protocol mechanisms encompass the communication and coordination techniques that enable structured context exchange between AI systems and external data sources \cite{hou2025model, ehtesham2025surveyagentinteroperabilityprotocols}. Building on standardized client–server architectures, they define how contextual information flows between distributed agents \cite{krishnan2025advancing}. Unlike traditional transports that emphasize data delivery, MCP mechanisms focus on semantic context representation, capability negotiation, and resource discovery across heterogeneous tool ecosystems \cite{ray2025survey}. Prior studies suggest this design can improve interoperability and real-time context sharing, while also raising concerns about scalability, overhead, and maintaining robust security boundaries across diverse applications.

\subsection{Design Philosophy and Session Model}

The foundational mechanism of MCP is a persistent client–server architecture intended to support long-lived context-exchange sessions between autonomous agents (or agent hosts) and external data sources \cite{hou2025model, krishnan2025advancing}. In contrast to stateless REST APIs, it maintains bidirectional connections that allow real-time updates and continuous tool interactions \cite{ehtesham2025surveyagentinteroperabilityprotocols}. This persistence can improve efficiency in multi-step interactions but increases resource consumption at scale. For adaptive systems, an MCP session can be interpreted as an adaptive state container that holds shared context, tool capabilities, and policy-relevant signals over time, providing a locus at which closed-loop behavior can be observed and evaluated.

The client–server model relies on a trusted host process that coordinates multiple client lifecycles while enforcing privacy controls and capability boundaries \cite{ray2025survey}. Session establishment typically follows capability negotiation, where clients and servers exchange capability advertisements during initialization \cite{hou2025model}. MCP also permits dynamic capability updates, enabling servers to advertise new tools or revoke access to existing ones without terminating the session \cite{krishnan2025advancing}. Viewed through an autonomy lens, this negotiation and update process implements dynamic action-space discovery: agents learn at run time which sensing, reasoning, and actuation options are available, and can adjust their policy spaces as tools appear, disappear, or change permissions. Security mechanisms such as role-based access control and OAuth 2.1-compliant authentication provide fine-grained permission management \cite{ray2025survey}, though reliance on external identity infrastructures may limit applicability in lightweight or decentralized deployments.

In highly dynamic transport scenarios—such as vehicles moving between roadside units or across cellular coverage boundaries—persistent MCP sessions will routinely encounter IP address changes, path breaks, and short-lived disconnections. Open questions include whether to re-bind sessions rapidly to new transport paths, shard sessions at edge nodes, or use higher-level idempotent reconnection patterns that trade strict continuity for safety and consistency. These design choices require systematic evaluation with the MCP-specific metrics outlined in our evaluation section before MCP can be trusted in safety-critical transport deployments.

\subsection{Messaging, Discovery, and Quality of Service}

MCP implements context exchange through a JSON-RPC 2.0-based messaging protocol (see Box~\ref{box:json}) that defines structured communication patterns \cite{ehtesham2025surveyagentinteroperabilityprotocols, ray2025survey}. Four message types—requests, results, errors, and notifications \cite{hou2025model}—provide predictability and compatibility with existing systems, though JSON-RPC’s verbosity and text encoding can create bandwidth inefficiencies in data-intensive scenarios. Standardized schemas embed metadata for provenance, temporal validity, and uncertainty quantification \cite{krishnan2025advancing, ray2025survey}, enabling informed decisions but increasing message size and potentially reducing throughput under high-frequency exchange. Versioned schema evolution addresses backward compatibility, while message routing and reliability mechanisms such as intelligent dispatching, retry with backoff, and compression improve resilience \cite{hou2025model, alla2025scalable}, at the cost of additional protocol overhead.

MCP’s resource discovery mechanisms enable dynamic identification of tools and data sources through standardized capability advertisement \cite{krishnan2025advancing}. These rely on semantic ontologies to match client needs to server capabilities \cite{hou2025model}, fostering interoperability but depending heavily on accurate and consistent descriptions. Tool orchestration provides coordination primitives for sequential, parallel, and conditional workflows \cite{ray2025survey, ehtesham2025surveyagentinteroperabilityprotocols}, making complex workflows manageable while introducing overhead, dependency bottlenecks, and error-propagation risks. Dynamic resource binding, graceful degradation, load balancing, and circuit breaker patterns \cite{krishnan2025advancing, hou2025model, ray2025survey} can improve resilience, but also increase implementation complexity and often still require human intervention when orchestrations fail.

MCP incorporates semantic awareness into contextual message routing \cite{hou2025model}, enabling prioritization of critical messages and optimization of delivery paths \cite{krishnan2025advancing}, which is particularly attractive for safety-critical or latency-sensitive applications. Semantic routing, however, requires extra computation for context inspection and may increase processing delay in high-throughput environments. QoS mechanisms provide differentiated service levels from best-effort delivery to bounded-latency guarantees \cite{ray2025survey, ehtesham2025surveyagentinteroperabilityprotocols}, but strict enforcement can strain resources during peak demand. Adaptive routing and context aggregation, such as batching, improve bandwidth efficiency \cite{alla2025scalable} yet risk delayed delivery if thresholds are not tuned carefully. Taken together, these mechanisms illustrate MCP’s ambition to unify diverse approaches to context exchange, while existing studies also caution about protocol complexity, overhead, and dependence on semantic infrastructures that may not scale across all AI-driven domains.

\section{MCP in Context-Aware Transport Systems}
\label{sec:context}

The transport-focused case study presented here is intentionally prospective. Existing reports in the MCP literature emphasize small-scale prototypes and experimental implementations, with limited evidence of large-scale, production-grade deployment in transport systems. Accordingly, we treat transport as a representative safety-critical domain for autonomous and adaptive systems and use it to illustrate how MCP’s session, context, and capability abstractions can structure distributed sense–reason–act loops spanning vehicles, infrastructure, and cloud services.

At a high level, the extended case study (see Appendix~\ref{app:context}) elaborates three dimensions: (i) environmental and application context, where MCP standardizes perception and intent signals across heterogeneous stacks; (ii) network state awareness, where MCP binds SDN/V2X telemetry to adaptation decisions; and (iii) integration with multiple communication paradigms, where MCP-inspired mechanisms complement IoT, cloud-native, and multi-modal transport platforms. These examples ground the autonomy-centric analysis in concrete scenarios while the main text emphasizes MCP’s general role as an adaptation substrate and coordination layer for distributed autonomous systems.

\section{Performance Analysis and Evaluation
}
\label{sec:evaluation}

 \subsection{Evaluation Methodologies and Metrics}
Evaluating adaptive transport systems requires methodologies that capture behavior across diverse conditions, applications, and adaptation strategies \cite{atkin2003evaluation, gama2014survey}. Traditional network metrics such as throughput, latency, and packet loss provide baselines but do not by themselves reflect the benefits or risks of adaptation \cite{atkin2003evaluation, zou2006adaptive}, so evaluation must also consider responsiveness, stability, and overhead \cite{atkin2003evaluation, anvari2023adaptive}. Simulation-based studies offer controlled environments for exploring parameters and scenarios that are difficult to reproduce in testbeds \cite{atkin2003evaluation, gama2014survey}, provided that models capture network and mobility complexity \cite{atkin2003evaluation, laicontext}. Testbed and pilot deployments validate simulation results and expose practical challenges \cite{atkin2003evaluation, zou2006adaptive} but are constrained by cost and scale \cite{atkin2003evaluation, gama2014survey}. Analytical modeling and trace-driven, data-centric evaluation complement these approaches by providing theoretical insight and realism, respectively, at the cost of simplifying assumptions or data limitations.

Performance metrics for adaptive transport systems must capture both traditional network characteristics and adaptation-specific behaviors \cite{atkin2003evaluation, zou2006adaptive}. Throughput, latency, and reliability provide baseline assessment but must be supplemented with metrics such as convergence time, stability, overhead, and robustness to changing conditions \cite{atkin2003evaluation, gama2014survey, anvari2023adaptive}. Energy efficiency is increasingly important for mobile and IoT applications \cite{pejovic2011context, talkingiot2025protocolwars}; adaptive decisions can significantly affect consumption, making energy-aware evaluation essential \cite{heinzelman1999adaptive, pejovic2011context}. Quality of Experience (QoE) metrics provide a user-centric view of adaptation by capturing end-user impact and application-specific preferences \cite{linqing2023dynamic, jevinger2023context, pejovic2011context, atkin2003evaluation}, and the development of standardized QoE measures for adaptive transport remains an active research area \cite{linqing2023dynamic, gama2014survey}.

 \subsection{Autonomous Adaptation and Comparative Analysis}
For MCP-enabled autonomous and adaptive systems, evaluation must go beyond link-level and flow-level performance to characterize properties of the closed-loop behavior induced by context exchange and capability negotiation. In addition to throughput, latency, and reliability, relevant metrics include adaptation latency (how quickly agents update policies or actions after context changes), policy stability (whether adaptation converges or oscillates), and coordination consistency (the fraction of decisions that remain jointly feasible across interacting agents and control loops).

MCP-specific evaluations should also quantify decision conflict rates—how often independently adapted agents propose incompatible actions given shared context—and recovery under partial observability, where sessions may experience missing, delayed, or stale context. These autonomy-centric metrics can be layered on top of existing methodologies from adaptive transport evaluation: simulations to stress-test multi-agent scenarios, testbeds to observe real-world coordination under mobility and disconnections, and trace-driven analyses to replay logged MCP sessions. By reporting such metrics alongside traditional network and QoE indicators, studies can assess whether MCP’s additional protocol complexity measurably improves the quality, safety, and robustness of autonomous adaptation relative to non-MCP baselines.

Comparative analysis of adaptive transport systems requires careful selection of scenarios, baselines, and metrics \cite{atkin2003evaluation, gama2014survey}. Fair comparison demands that systems be evaluated under identical conditions with baselines representing current best practice \cite{atkin2003evaluation, zou2006adaptive}. Because approaches are diverse, comprehensive frameworks are needed \cite{anvari2023adaptive, gama2014survey}. Analyzing adaptation overhead is critical for judging practical viability: computational, communication, and storage costs must be weighed against performance gains \cite{atkin2003evaluation, pejovic2011context}, and systems that achieve modest improvements at high overhead may be unsuitable for deployment \cite{atkin2003evaluation, anvari2023adaptive}. Robustness analysis examines how adaptive systems maintain performance under adverse conditions, failures, and attacks \cite{zou2006adaptive, hou2025model}, recognizing that increased complexity can introduce new vulnerabilities \cite{zou2006adaptive}. Security analyses of MCP systems highlight the need to treat security as integral to adaptive design \cite{hou2025model}. Table~\ref{tab:evaluation_methods} summarizes the main evaluation methodologies, associated metrics, and limitations.

\begin{table}[htbp]
\centering
\small
\caption{Evaluation methodologies and metrics in adaptive transport systems.}
\label{tab:evaluation_methods}
\renewcommand{\arraystretch}{1.2}
{\fontsize{8}{9}\selectfont
\begin{tabularx}{\textwidth}{L{2cm}L{3cm}L{2.8cm}L{3cm}X}
\toprule
\rowcolor{blue!6}
\textbf{Methodology} & \textbf{Description / Examples} & \textbf{Typical Metrics} & \textbf{Limitations} & \textbf{References} \\
\midrule
Simulation-Based Studies &
Microscopic or mesoscopic traffic simulators (e.g., SUMO, VISSIM). &
Travel time, delay, throughput, emissions, energy efficiency. &
Simplified assumptions may not capture real-world variability. &
\cite{hossain2024vissim, maadi2022real} \\
\midrule
Testbed / Pilot Deployments &
Experimental or small-scale real-world setups. &
Reliability, latency, packet loss, and coordination success rate. &
High cost; limited scalability for large-scale validation. &
\cite{cai2024adaptive, humdan2025} \\
\midrule
Analytical Modeling &
Mathematical or theoretical protocol models. &
Stability, convergence, scalability bounds. &
Difficult to incorporate stochasticity and context variability. &
\cite{velayutham2005transport, he2005survey} \\
\midrule
Trace-Driven / Data-Centric Evaluation &
Use of sensor logs, field-collected data, or traffic traces. &
Prediction accuracy, generalization, robustness. &
Data quality and availability constraints; reproducibility challenges. &
\cite{du2024survey, li2015context} \\
\midrule
Hybrid Approaches &
A combination of simulation, modeling, and real data. &
Multi-metric evaluation across both network and transport performance. &
Complex setup; results may be harder to interpret consistently. &
\cite{ray2025survey, alanazi2024framework} \\
\bottomrule
\end{tabularx}
}
\end{table}

In summary, Table~\ref{tab:evaluation_methods} highlights that no single methodology provides comprehensive coverage: simulations offer scalability but risk oversimplification, testbeds provide realism at higher cost, and analytical or trace-driven studies contribute rigor while facing reproducibility constraints. For MCP-enabled architectures, these methodologies must additionally capture protocol-specific dimensions such as session churn and recovery time under mobility, context freshness and propagation delay along MCP channels, capability discovery latency and failure rates, and schema evolution or validation overhead. Hybrid evaluation campaigns that combine traffic simulators, MCP-aware testbeds, and data-centric analysis are therefore particularly important for quantifying the end-to-end impact of MCP mechanisms on adaptive transport performance.

\section{Challenges and Open Problems
}
\label{sec:challenges}

Despite significant advances in adaptive transport systems and MCP, numerous challenges and unresolved research questions persist. This section highlights the most pressing issues that constrain scalability, security, and interoperability.

\subsection{Scalability and Technical Complexity}

As the number of connected devices, sensors, and autonomous vehicles grows, adaptive transport systems must process increasing volumes of real-time data and contextual information, placing substantial load on network and computing infrastructure \cite{itsa2024intelligent, xenatech2024its}. Meeting these demands while integrating AI, IoT, and edge–cloud infrastructures requires scalable architectures, specialized expertise, and sustained investment across deployment, upgrades, and operations; Table~\ref{tab:challenges_open_problems} summarizes the main scalability and complexity issues.

\subsection{Security and Privacy}

Greater interconnectedness and dependence on data expand the attack surface of adaptive transport systems, from roadside sensors and in-vehicle networks to cloud-based services, increasing exposure to malware, unauthorized access, and denial-of-service attacks \cite{itsdigest2023gao,geldner2024dtls}. At the same time, pervasive collection of sensitive data such as locations, travel patterns, and identifiers raises demanding privacy and compliance requirements \cite{waqar2023evaluation}; addressing these security and privacy challenges, outlined in Table~\ref{tab:challenges_open_problems}, is essential for safe MCP adoption.
\subsection{Standardization and Interoperability}

Persistent fragmentation of architectures and communication protocols across agencies and vendors \cite{agbaje2022survey,concas2024mobility,itsdigest2023gao} continues to hinder large-scale deployment of adaptive transport systems, especially in Internet of Vehicles settings where heterogeneous vehicles, infrastructure, and cloud services must coordinate dynamically. Beyond syntactic compatibility, Table~\ref{tab:challenges_open_problems} highlights the need for semantic interoperability and shared governance so that systems can correctly interpret exchanged data and avoid unsafe or inefficient decisions.

\begin{table}[htbp]
\centering
\small
\caption{Summary of Challenges and Open Problems in Adaptive Transport Systems.}
\label{tab:challenges_open_problems}
\renewcommand{\arraystretch}{1.2}
{\fontsize{8}{9}\selectfont
\begin{tabularx}{\textwidth}{L{2cm} L{3cm} X}
\toprule
\rowcolor{blue!6}
\textbf{Category} & \textbf{Sub-Issue} & \textbf{Description} \\
\midrule
\textbf{Scalability \& Technical Complexity} & Data Volume \& System Load & Rapid growth in connected devices creates data surges requiring robust real-time processing, storage, and transmission. \\
& Technology Integration & Incorporating AI, IoT, and edge computing increases system design complexity, demanding high technical expertise. \\
& Initial Investment \& Upgrades & High upfront costs for hardware/software, risk of technology obsolescence, and ongoing upgrade requirements. \\
& Operations \& Maintenance & Complex, aging systems require dedicated maintenance resources to ensure uptime and consistent performance. \\
\midrule
\textbf{Security \& Privacy} & Cybersecurity Threats & Interconnected systems are vulnerable to attacks like malware, unauthorized access, and DDoS, often lacking sufficient protections. \\
& Data Privacy & Collection of sensitive user data (e.g., location, travel habits) raises concerns over compliance and safeguarding. \\
\midrule
\textbf{Standardization \& Interoperability} & Fragmented Systems & Diverse, proprietary systems across vendors and agencies hinder communication and cross-platform integration. \\
& Lack of Uniform Standards & Absence of standardized data exchange and protocols limit system scalability and cohesion, especially in the Internet of Vehicles. \\
& Semantic Interoperability & Systems must agree not just on format, but on the meaning of shared data to ensure coordinated and accurate decision-making. \\
\bottomrule
\end{tabularx}
}
\end{table}

In summary, realizing adaptive, resilient, and intelligent transport systems will require addressing these interconnected challenges through continued research, interdisciplinary collaboration, supportive policy, and sustained investment from both public and private sectors.

\section{Future Research Directions
}
\label{sec:future}

Drawing on the preceding analysis, we prioritize three near-term MCP-related research questions for adaptive transport systems: (i) how to design and evaluate MCP-enabled architectures under realistic mobility and disconnection patterns; (ii) how to integrate MCP with existing ITS, IoT, and middleware standards without duplicating or undermining their functionality; and (iii) how to govern schemas, security boundaries, and capability negotiation when MCP is deployed in safety-critical transport settings. The subsections below expand on these directions while also outlining longer-term opportunities.

\subsection{AI-Driven and Edge-Centric Adaptation}
The integration of artificial intelligence and machine learning into adaptive transport systems is a central research direction \cite{anvari2023adaptive, laicontext}. AI-driven adaptation can outperform rule-based approaches by learning strategies from data and adjusting automatically to changing conditions \cite{gama2014survey, anvari2023adaptive}, but it raises open questions around online learning under mobility, safe reinforcement learning, and federated training that preserves privacy while leveraging cross-system data \cite{nsrg2019adaptive, sung2023decentralized}. For MCP-enabled architectures, a concrete challenge is to represent learning state, exploration constraints, and safety margins as first-class context objects so that convergence, regret, and robustness under disruptions can be evaluated against non-MCP baselines.

Edge computing offers complementary opportunities by enabling distributed processing of context information and adaptation decisions \cite{bhardwaj2020adaptive, sung2023decentralized}. Edge resources can reduce adaptation latency and improve responsiveness \cite{bhardwaj2020adaptive, alanazi2024framework}, but coordination across heterogeneous edge nodes requires scalable consensus and orchestration mechanisms \cite{kumar2003dynamic, govloop2020adaptivenetwork}. For MCP, this points to research on where to place servers and hosts along the edge–cloud continuum, how to route context sessions across edge domains, and how to balance edge-local aggregation against end-to-end consistency in realistic mobility settings.

\subsection{Long-Term Directions: Quantum Communication and Coordination}
At present there is no concrete integration between MCP and quantum communication in transport deployments; we therefore treat this topic as a long-term outlook on how MCP’s context and session abstractions might eventually coexist with quantum-safe networking and cryptography, including post-quantum key management and migration strategies that avoid disrupting safety-critical operation.

For autonomous transport agents, MCP’s primary near-term role is to act as a structured context- and tool-exchange layer, while the underlying coordination mechanisms (e.g., auction-based, game-theoretic, or blockchain-based) remain largely orthogonal and domain dependent. This raises questions about how MCP should expose negotiation state, safety constraints, and accountability information to existing coordination frameworks without introducing new consensus protocols; given the costs and governance challenges of distributed ledgers, we regard blockchain-based coordination in MCP-enabled transport as a speculative, longer-term direction.

\section{Conclusions
}
\label{sec:conclusion}

This survey has examined unifying models for adaptive transport systems through the lens of autonomous and adaptive behavior, with particular emphasis on the Model Context Protocol (MCP) as an adaptation substrate. By integrating nearly three decades of work on adaptive protocols, context-aware frameworks, and integration architectures, and by reframing them as loci of adaptation in distributed autonomy, we have outlined how MCP’s session, context, and capability abstractions could structure sense–reason–act loops that span vehicles, infrastructure, and cloud services. Because MCP deployments in transport remain limited, our claims about scalability, interoperability benefits, and suitability for safety-critical settings should be read as conceptually grounded hypotheses that require empirical validation using the autonomy-centric evaluation agenda developed in this article.

Our analysis indicates that while individual components of adaptive transport systems are increasingly sophisticated, the field still lacks general abstractions for exposing and coordinating adaptation state across agents and layers. MCP contributes such an abstraction by externalizing context and capabilities, but it also inherits challenges around scalability, security, governance, and standardization. We therefore view MCP-enabled transport systems as a representative domain for a broader class of autonomous and cyber-physical systems, including robotics fleets, cyber–physical infrastructure, edge AI deployments, and multi-agent systems, where similar tensions between local autonomy and global coordination arise.

Looking ahead, the success of MCP-inspired architectures will depend on whether they can measurably improve the quality, safety, and robustness of autonomous adaptation while coexisting with established protocols and middleware. Realistic evaluations that capture adaptation latency, policy stability, coordination consistency, and recovery under partial observability, together with cross-domain case studies beyond transport, will be crucial. By articulating MCP as a coordination layer for autonomous control loops and by grounding this view in existing transport literature, we aim to provide a foundation for future work that designs, implements, and rigorously evaluates autonomous systems built on shared context substrates, in transport and in adjacent domains.

\bibliographystyle{ACM-Reference-Format}
\bibliography{references-checked}

\newpage
\appendix

\section{MCP in Context-Aware Transport Systems}
\label{app:context}

This section is deliberately prospective. The published evidence on MCP deployments in transport is currently dominated by small-scale prototypes rather than mature, large-scale operational systems. Accordingly, we present the scenarios below as conceptual instantiations of the mechanisms in \S\ref{sec:protocols}, emphasizing MCP as an adaptation substrate for distributed autonomous agents and making explicit where examples conform to the current specification or rely on additional assumptions.

Context-aware frameworks form the intelligence layer of adaptive transport systems, implementing sense–reason–act loops that allow vehicles, infrastructure controllers, and backend services to perceive the environment, infer state, and trigger coordinated actions. Within MCP, these frameworks can act as both producers and consumers of structured context, exchanging semantically rich, uncertainty-aware representations through a standardized interface. MCP has been proposed as a unifying layer for diverse context flows across vehicles, infrastructure, edge nodes, and control centers, enabling agents to share world models, policies, and constraints rather than only raw data. While this aligns with longstanding interoperability challenges, questions remain about scalability, overhead, and standardization costs. We therefore focus on three dimensions of context awareness—environmental context processing, application context recognition, and network state awareness—each as a potential locus where MCP could support cross-agent adaptation but also where limitations and open research questions emerge.

\subsection{Environmental and Application Context}

Environmental context processing in intelligent transportation follows a loop of acquisition, fusion, semantic interpretation, and exposure, in which heterogeneous sensor inputs (LiDAR, cameras, radar, IMU, GNSS, HD maps) are transformed into probabilistic spatial states for planning and control~\cite{baldauf2007survey, hong2009context, dey2001understanding}. Deep multimodal fusion generally outperforms unimodal perception in adverse conditions such as fog, rain, or night driving, with comparative reviews distinguishing data-, feature-, and decision-level strategies~\cite{fayyad2020deep, tang2023comparative, zhang2023perception}. Cooperative perception extends this by enabling vehicles and roadside infrastructure to exchange object lists or BEV features, improving detection at occluded intersections~\cite{bai2024survey, ji2024toward}. Edge platforms such as MEC and VEC often host intensive fusion stages, reducing latency for perception sharing near RSUs~\cite{brehon2022mobile, ahmed2022survey, yusuf2024vehicle}. MCP could standardize how these perception outputs are packaged and exchanged, representing environmental state updates as structured, uncertainty-aware context objects with semantic labels, timeliness metadata, and provenance tracking so that perception data is more easily consumable across heterogeneous stacks—for example, RSUs publishing occupancy grids or lane geometry with confidence scores for nearby vehicles. At the same time, standardization may constrain innovation in perception pipelines, transmitting rich fused context at scale can strain bandwidth and processing, and the reliability of shared context depends on source trustworthiness that MCP alone cannot guarantee.

Application context recognition translates user state, agent intent, and task parameters into adaptation signals for navigation, MaaS, human–machine interaction, and safety logic. Examples include transportation-mode detection from smartphone sensors~\cite{tang2024feature, kamalian2022survey, sadeghian2021review}, driver monitoring via fatigue or distraction detection~\cite{li2021survey, nees2022mental}, and pedestrian trajectory and intention prediction for vulnerable road users~\cite{sharma2022pedestrian, zhou2024pedestrian}. Multi-task and intention-aware models typically outperform single-task baselines, enabling earlier hazard detection under distribution shifts~\cite{sharma2022pedestrian, gomes2025comprehensive, xia2024survey}. In MaaS, multi-source behavioral context such as time, weather, and social data improves demand forecasting and recommendations~\cite{servizi2021transport}. MCP offers one mechanism for structuring and disseminating such context across heterogeneous systems: events like “driver distracted,” “pedestrian likely to cross in 3s,” or “group prefers low-transfer route” can be published as machine-readable, confidence-tagged objects, allowing automated vehicle decision modules to consume third-party driver-monitoring outputs or MaaS platforms to integrate risk-aware pedestrian forecasts. However, interoperability depends on shared semantic definitions and schema evolution, differences in labeling or uncertainty thresholds can undermine consistency, and protecting behavioral data raises privacy and governance concerns, so MCP deployment in application-context domains requires careful attention to interpretability, standardization, and user trust.

\subsection{Network State Awareness}

Network state awareness provides real-time telemetry on topology, load, latency, and link quality across SDN-enabled V2X and MEC infrastructures, helping keep sensing, computation, and control within QoS and SLA targets~\cite{varma2023comprehensive, tan2021band, sohail2023routing}. Work on VANET routing, GNN-based traffic prediction~\cite{jiang2022graph, ahmed2024enhancement}, and SDN orchestration~\cite{mekki2022software, abdullah2023edge} shows the value of coupling application urgency with network state to optimize resource allocation—for example, placing compute at RSUs to reduce latency for cooperative perception or using prediction-driven routing to pre-warm paths for safety-critical updates~\cite{mendiboure2019edge}. MCP could bind network telemetry to adaptive behavior by standardizing metrics such as slice capacity, link delay, or congestion forecasts. An orchestrator might publish “link degradation between RSU-23 and RSU-24, expected latency +30 ms,” prompting applications to adjust fusion pipelines or send rates. Such alignment can improve resilience but depends on accurate, timely telemetry; continuous reporting may burden constrained networks, and complex semantic QoS definitions may hinder adoption. Prioritization across stakeholders (e.g., private fleets vs. public safety systems) also raises fairness and governance issues. From a system-dynamics perspective, highly dynamic topologies and frequent disconnections in V2X and MEC deployments make it essential to study how MCP sessions and context streams behave under route changes, handoffs, and temporary outages.

\subsection{Illustrative MCP-Enabled Transport Scenarios}

Given that MCP remains an emerging framework, documented transport-domain deployments are presently concentrated at the prototype and experimental stage rather than at large-scale operational maturity. Nevertheless, its abstractions can be systematically mapped onto transport scenarios to clarify potential usage and architectural implications. In a cooperative-perception corridor, roadside units (RSUs) and environmental sensors could expose lane-level occupancy and hazard summaries via an MCP server, while vehicles operate as MCP clients that subscribe to structured context objects and publish localized observations, enabling cross-vendor perception without reliance on a single proprietary stack. Within a Mobility-as-a-Service (MaaS) platform, MCP could mediate between journey planners, ticketing systems, and pricing engines, standardizing access to demand forecasts, disruption alerts, and user preferences while established ITS standards continue to govern low-level signaling and control. Similarly, a traffic-operations control center could treat incident-management tools, predictive analytics pipelines, and digital-twin simulators as MCP tools bound to a shared context graph, thereby simplifying orchestration of scenario analysis and coordinated interventions. Although conceptual, these scenarios provide architectural blueprints for simulation studies and controlled testbeds as MCP implementations progress toward broader deployment.

\subsection{MCP Unification: Schemas, Tools, and Multi-Agent Coordination}

MCP Unification Models represent architectural frameworks and standardization mechanisms that integrate diverse AI tools, context sources, and applications into coherent, interoperable ecosystems \cite{hou2025model, krishnan2025advancing}. They address persistent semantic and syntactic heterogeneity in AI integration, where tools, data sources, and reasoning engines differ in interfaces, formats, and operational paradigms \cite{ehtesham2025surveyagentinteroperabilityprotocols}. Unlike point-to-point adapters or monolithic middleware, MCP proposes a standardized protocol layer that abstracts integration complexity while preserving system-specific capabilities \cite{ray2025survey}. We therefore review three dimensions of MCP unification: context standardization, AI tool integration, and multi-agent coordination.

Context standardization in MCP targets the difficulty of representing heterogeneous contextual information in machine-interpretable, interoperable formats \cite{hou2025model}. Traditional systems often exhibit semantic fragmentation, with incompatible representations for similar concepts \cite{contextmodel2016, 5483753}. MCP addresses this through schema definitions and ontological frameworks intended to provide consistency across domains \cite{ray2025survey}. Separating syntactic representation (for example JSON-RPC message formats) from semantic meaning (ontological relationships) allows extensibility while maintaining protocol-level compatibility \cite{krishnan2025advancing, ehtesham2025surveyagentinteroperabilityprotocols}. Versioned schema evolution with backward compatibility supports changing requirements in long-lived deployments \cite{hou2025model, ray2025survey}, while provenance and constraint checking help maintain semantic consistency and data quality \cite{5483753}. However, governance may become a bottleneck: maintaining alignment across contributors is difficult, version fragmentation can arise if consensus over changes is slow, and strict validation may reject unconventional but useful contextual signals, trading off interoperability against adaptability.

AI tool integration through MCP unifies diverse reasoning systems, data-processing tools, and AI services into shared workflows \cite{krishnan2025advancing}. Capability abstraction and negotiation let tools advertise functions in standardized ways \cite{hou2025model}, reducing friction from incompatible interfaces \cite{ray2025survey}. Dynamic discovery and negotiation support flexible composition while respecting tool limitations \cite{ehtesham2025surveyagentinteroperabilityprotocols}. Workflow orchestration covers sequential, parallel, and conditional execution \cite{ray2025survey}, supported by primitives for data flow, error propagation, and transactional consistency \cite{hou2025model}. Optimizations such as token-based referencing and distributed caching reduce payloads and communication overhead \cite{alla2025scalable, ray2025survey}, while streaming coordination mitigates latency. Error-handling techniques—circuit breakers, retries, and compensation—enhance robustness \cite{krishnan2025advancing, hou2025model}. At the same time, orchestration complexity can grow quickly as workflows span heterogeneous tools, complicating debugging and consistent semantics across vendors.

Multi-agent coordination frameworks in MCP support distributed AI collaboration on tasks that exceed the scope of individual agents \cite{krishnan2025advancing}. Coordination models range from hierarchical delegation and peer-to-peer negotiation to consensus-driven decision making \cite{ray2025survey, ehtesham2025surveyagentinteroperabilityprotocols}, echoing long-standing approaches in multi-agent systems \cite{du2024survey}. MCP’s contribution lies in standardized communication, conflict resolution, and consensus mechanisms tailored for AI tool ecosystems \cite{hou2025model}. Context sharing across agents must balance freshness with efficiency, so MCP employs distributed context-management and reconciliation strategies to maintain eventual consistency \cite{alla2025scalable, krishnan2025advancing}. These enable collaboration but introduce complexity in maintaining timeliness under high churn. Conflict resolution via timestamping or semantic reconciliation can reduce inconsistencies but may not scale under highly asynchronous conditions. Collaborative intelligence is framed as the outcome of integrating agents specializing in NLP, vision, reasoning, or domain expertise \cite{ray2025survey, hou2025model}, with mechanisms such as attention and contribution weighting producing coherent outputs from diverse inputs. However, coordination overhead can erode performance benefits when communication dominates computation, and security and trust remain challenging: authentication, authorization, and audit mechanisms set collaboration boundaries \cite{hou2025model, ray2025survey}, but consistent enforcement across distributed, possibly adversarial participants is difficult and must be complemented by governance addressing incentives, fairness, and accountability.

\subsection{Integrating Multiple Communication Paradigms}
\label{sec:integration}

Transport system integration requires bridging multiple communication paradigms, device capabilities, and cloud infrastructures in a way that maintains interoperability, performance, and security. Recent work positions the MCP as a useful conceptual parallel for these efforts, given its focus on structured context exchange, persistent session management, and schema-based interoperability \cite{hou2025model, ray2025survey}. However, transport systems pose unique constraints—resource limitations in IoT, dynamic handoffs in multi-modal environments, and elasticity in cloud-native infrastructures—that highlight both opportunities and open challenges for applying MCP-inspired principles. Table~\ref{tab:integration} summarizes the key integration domains and their corresponding mechanisms, challenges, and MCP analogies.

\begin{table}[htbp]
\centering
\small
\caption{Comparison of integration dimensions in transport systems and MCP-inspired mechanisms.}
\label{tab:integration}
\renewcommand{\arraystretch}{1.2}
{\fontsize{8}{9}\selectfont
\begin{tabularx}{\textwidth}{L{2cm}L{3cm}L{3cm}XX}
\toprule
\rowcolor{blue!6}
\textbf{Integration Dimension} & \textbf{Key Challenges} & \textbf{MCP-Inspired Mechanisms} & \textbf{Strengths} & \textbf{Limitations} \\
\midrule
Multi-modal Transportation & 
Seamless switching across heterogeneous networks (wired, wireless, vehicular, satellite); preserving QoE across client–server, publish–subscribe, and peer-to-peer paradigms \cite{cordis101117675, russo2012unifying}. &
Standardized APIs for abstraction \cite{elatec2025smartmoves}; token-based referencing for efficient mode switching \cite{alla2025scalable}; lifecycle management for trips akin to context sessions \cite{hou2025model}. &
Flexibility through abstraction; reduces integration complexity; enables adaptive switching across modes. &
Scalability and vendor adoption challenges; risk of overhead during frequent transitions; requires consensus on APIs. \\
\midrule
IoT Integration &
Constrained devices, intermittent connectivity, protocol fragmentation \cite{talkingiot2025protocolwars}. &
Lightweight persistent sessions \cite{patil2025inside}; schema-flexible payloads; OTA updates and device onboarding; edge caching with tokenized contexts \cite{alla2025scalable}. &
Supports constrained devices; improves reliability during disconnections; scalable OTA management. &
Security vulnerabilities in heterogeneous fleets; semantic interoperability difficulties; added complexity at gateways. \\
\midrule
Cloud-Native Protocols &
Elastic scaling, serverless lifecycles, and heterogeneous edge–cloud topologies \cite{alanazi2024framework}. &
Session-oriented authentication (OAuth 2.1) \cite{ray2025survey}; QUIC/HTTP3-based low-latency transport \cite{zhu2025unifiedworldmodelscoupling}; observability hooks for adaptive routing and fault isolation \cite{jiang2024survey}. &
High performance under elasticity; strong alignment with microservice and zero-trust security models. &
Cold-start latency in serverless contexts; fairness across tenants; lack of standardization in APIs. \\
\bottomrule
\end{tabularx}
}
\end{table}

\vspace{10pt}

\noindent \textbf{Multi-modal Transportation}: Multi-modal systems increasingly span heterogeneous network types and paradigms, including client–server, publish–subscribe, and peer-to-peer \cite{russo2012unifying, elatec2025smartmoves}. Ensuring seamless transitions between these fabrics while preserving service continuity remains a central challenge \cite{govloop2020adaptivenetwork}. As shown in Table~\ref{tab:integration}, abstraction layers and unified APIs parallel MCP’s standardized interfaces, decoupling application workflows from underlying transport specifics \cite{hou2025model, singh2025survey}. MCP-inspired features like token-based referencing \cite{alla2025scalable} and lifecycle-aware context management \cite{hou2025model} illustrate how transport APIs might reduce signaling overhead during mode switching, though issues of scalability and vendor adoption remain open.

\vspace{10pt}
\noindent \textbf{IoT Integration}: IoT integration adds complexity due to device constraints, intermittent links, and protocol fragmentation \cite{talkingiot2025protocolwars}. MCP’s session model and schema-flexible payloads offer lessons for lightweight, persistent telemetry management in such environments \cite{patil2025inside}. Table~\ref{tab:integration} highlights parallels between MCP’s OTA lifecycle management and IoT fleet requirements such as secure onboarding, firmware updates, and buffering during disconnections. Optimizations like FastMCP’s token-based caching and Redis-backed persistence \cite{alla2025scalable} demonstrate potential scalability, though operational deployments still face security and semantic interoperability challenges.

\vspace{10pt}
\noindent \textbf{Cloud-Native Protocols}: Cloud-native protocols must adapt to ephemeral containers, serverless execution, and edge-cloud continuum architectures \cite{bhardwaj2020adaptive, alanazi2024framework}. Protocols like QUIC are increasingly adopted to reduce latency and manage multiplexed streams \cite{zhu2025unifiedworldmodelscoupling}, while service meshes impose requirements for adaptive routing, encryption, and observability. As summarized in Table~\ref{tab:integration}, MCP’s emphasis on capability negotiation, session management, and secure delegation resonates with these demands \cite{hou2025model, ray2025survey}. Yet the cloud-native domain raises unique concerns about cold-start latency, multi-tenant fairness, and standardization gaps, which remain unsolved despite architectural parallels with MCP.

\end{document}